\documentclass[conference]{IEEEtran}
\IEEEoverridecommandlockouts

\usepackage{cite}
\usepackage{amsmath,amssymb,amsfonts}

\usepackage{graphicx}
\usepackage{textcomp}
\usepackage{xcolor}

\usepackage{algorithm}
\usepackage{algpseudocode}
\usepackage{multirow}
\usepackage{makecell}
\usepackage{rotating}
\usepackage{adjustbox}
\usepackage{amsmath}
\usepackage{amssymb}
\usepackage{mathtools}
\usepackage{amsfonts}
\usepackage{relsize} 

\usepackage{rotating}
\usepackage{array}
\usepackage{makecell}
\usepackage{tabularx}

\usepackage{booktabs}
\usepackage{array}

\usepackage{caption} 

\def\BibTeX{{\rm B\kern-.05em{\sc i\kern-.025em b}\kern-.08em
    T\kern-.1667em\lower.7ex\hbox{E}\kern-.125emX}}
  
\begin{document}

\title{OLR-WA: Online Weighted Average Linear Regression in Multivariate Data Streams}

\author{\IEEEauthorblockN{1\textsuperscript{st} Mohammad Abu-Shaira}
\IEEEauthorblockA{\textit{Computer Science and Engineering} \\
\textit{Baylor University}\\
Waco, Texas, United States \\
mohammad\_abu-shaira1@baylor.edu}
\and
\IEEEauthorblockN{2\textsuperscript{nd} Alejandro Rodriguez}
\IEEEauthorblockA{\textit{Computer Science and Engineering} \\
\textit{Baylor University}\\
Waco, Texas, United States \\
alejandro\_rodriguez4@baylor.edu}
\and
\IEEEauthorblockN{3\textsuperscript{rd} Greg Speegle}
\IEEEauthorblockA{\textit{Computer Science and Engineering} \\
\textit{Baylor University}\\
Waco, Texas, United States\\
greg\_speegle@baylor.edu}
\and
{\,\hspace{5.6em}\,}
\and
\IEEEauthorblockN{4\textsuperscript{th} Victor Sheng}
\IEEEauthorblockA{\textit{Computer Science and Engineering} \\
\textit{Texas Tech University}\\
Lubbock, Texas, United States \\
victor.sheng@ttu.edu}
\and
\IEEEauthorblockN{5\textsuperscript{th} Ishfaq Ahmad}
\IEEEauthorblockA{\textit{Computer Science and Engineering} \\
\textit{University of Texas at Arlington}\\
Arlington, Texas, United States \\
iahmad@cse.uta.edu}
}

\maketitle

\begin{abstract}
Online learning updates models incrementally with new data, avoiding large storage requirements and costly model recalculations. In this paper, we introduce ``OLR-WA; OnLine Regression with Weighted Average'', a novel and versatile multivariate online linear regression model. We also investigate scenarios involving drift, where the underlying patterns in the data evolve over time, conduct convergence analysis, and compare our approach with existing online regression models. The results of OLR-WA demonstrate its ability to achieve performance comparable to the batch regression, while also showcasing comparable or superior performance when compared with other state-of-the-art online models, thus establishing its effectiveness. Moreover, OLR-WA exhibits exceptional performance in terms of rapid convergence, surpassing other online models with consistently achieving high $r^2$ values as a performance measure from the first iteration to the last iteration, even when initialized with minimal amount of data points, as little as 1\% to 10\% of the total data points. In addition to its ability to handle time-based (temporal drift) scenarios, remarkably, OLR-WA stands out as the only model capable of effectively managing confidence-based challenging scenarios. It achieves this by adopting a conservative approach in its updates, giving priority to older data points with higher confidence levels. In summary, OLR-WA's performance further solidifies its versatility and utility across different contexts, making it a valuable solution for online linear regression tasks.
\end{abstract}


\section{Introduction}
In Machine Learning, the conventional batch approach has some inherent limitations. Firstly, it assumes complete access to all data for every computation, making it impractical for scenarios with large or continuously changing datasets. Secondly, the batch model relies on the assumption that there are no time restrictions, which may not be feasible in real-time applications where timely predictions are required. Lastly, the batch model's rigidity in assuming a fixed and unchanging data distribution restricts its adaptability to situations where the data distribution evolves over time, rendering it less suitable for dynamic or non-stationary environments \cite{fontenla2013online}.

Over the past years, online models have emerged as an influential method for real-time predictive modeling, especially in dynamic and evolving environments. Our preliminary work, titled ``OnLine Regression Based on Weighted Average; OLR-WA'' \cite{olr_wa_23}, introduced a robust methodology that effectively handled 2-D and 3-D scenarios. However, the increasing demand for predictive models capable of effectively handling higher-dimensional data is paramount in addressing real-life complexities. To address this imperative, the present paper delves into the captivating realm of N-dimensional regression, offering a valuable technique to effectively tackle real-life complexities in dynamic and evolving environments.

This study is focused on the development of the multivariate OLR-WA model and its comparative performance evaluation against existing online regression models. We conduct a thorough comparative analysis, examining convergence properties and the model's ability to handle adversarial scenarios. 

Performance analysis on adversarial scenarios evaluates models in dynamic environments. This analysis uncovers vulnerabilities, informs effective strategies for handling changing data, and ensures system reliability and resilience \cite{papernot2016transferability}. To elucidate the concept of adversarial scenarios within the scope of our research, we introduce two distinct categories. The first category, denoted as ``Time-Based adversarial scenarios,'' entails the dynamic drift of data, resulting in a progressively divergent linear regression model over time. We investigate the models' capacity to adapt to these evolving patterns. In contrast, in the second category, referred to as ``Confidence-Based adversarial scenarios", even though data drift is present, the model is expected to stay conservative in its updates given its established confidence in its existing model.

This rigorous evaluation allows us to assess the effectiveness and competitiveness of our proposed approach. By bench-marking our model against established alternatives, we gain valuable insights into its strengths, weaknesses, and potential areas for enhancement. Such a comparative analysis serves to provide a deeper understanding of our model's capabilities and its relative performance in relation to other methodologies.

\section{Related Work}
This section explores online regression algorithms, evaluating their strengths, limitations, and real-world applicability, uncovering advancements and research directions in online linear regression.

\subsection{Online Regression using Stochastic Gradient Descent}
Stochastic Gradient Descent (SGD), commonly referred to as Online SGD \cite{ding2021efficient}\cite{tan2019online}. Although SGD processes only one data point per iteration, and this crucial difference enables it to efficiently handle large-scale datasets and significantly reduce the computational overhead, it necessitates access to all previously encountered data points, resulting in substantial memory requirements \cite{ruder2016overview}. Furthermore, SGD's faster computation comes at the cost of slower convergence \cite{johnson2013accelerating}. The selection of an appropriate learning rate is crucial in SGD's performance, as a too-high value may lead to overshooting, while a too-low value may cause slow convergence or sub-optimal solutions \cite{xu2017reinforcement}. Lastly, SGD's sensitivity to outliers may yield sub-optimal solutions due to their disproportionate influence on parameter updates \cite{shah2020choosing}. The loss function and the update rules are expressed through equations from \ref{eq:mse} to \ref{eq:stochastic_gd_derivative} \cite{goodfellow2016deep}.

{
\scriptsize
\begin{equation}
    \label{eq:mse}
        \text{MSE}_i = (y_i - \hat{y}_i)^2
\end{equation}
\vspace{-4.0mm}
\begin{equation}
    \label{eq:gd_update_rule}
    \begin{aligned}
    w &\leftarrow w - \eta \frac{\partial \text{MSE}_i}{\partial w} \quad ;
    b &\leftarrow b - \eta \frac{\partial \text{MSE}_i}{\partial b} 
    \end{aligned}
\end{equation}
\vspace{-4.0mm}
\begin{equation}
    \label{eq:stochastic_gd_derivative}
    \begin{aligned}
    \frac{\partial MSE_i}{\partial w} &= -2(y_i - \hat{y}_i) \cdot \mathbf{X}_i \quad ;
    \frac{\partial MSE_i}{\partial b} &= -2(y_i - \hat{y}_i)
    \end{aligned}
\end{equation}
}

The online stochastic method can be extended by involving $L_1$ and $L_2$ regularization, resulting in the online versions of lasso and ridge regression: online lasso regression (OLR) and online ridge regression (ORR), respectively. They both add a regularization term to the MSE loss that helps reduce overfitting by penalizing the model's complexity, resulting in more robust and generalized predictions \cite{tibshirani1996regression} \cite{abu-mustafa2012}. However, these methods require tuning the regularization parameter ($\lambda$) to control the trade-off between fitting the data and regularization. Selecting an optimal value for $\lambda$ can be challenging and may require experimentation or cross-validation \cite{hastie2009elements}\cite{chichignoud2016practical}\cite{stephenson2021can}\cite{mohri2018foundations}. 
The cost function with L1, and L2 regularization and its derivative represented by equations \ref{eq:lasso_mse}, \ref{eq:lass_w_b} and \ref{eq:mse_ridge}, \ref{eq:ridge_w_b}, respectively.

{
\scriptsize
\begin{equation}
    \label{eq:lasso_mse}
    \text{MSE}_i = (y_i - \hat{y}_i)^2 + \lambda \sum_{j=1}^{m} |w_j| = (y_i - \hat{y}_i)^2 + \lambda \|\mathbf{w}\|_1
\end{equation}
\vspace{-5.0mm}
\begin{equation}
    \label{eq:lass_w_b}
    \begin{aligned}
    \frac{\partial \text{MSE}_i}{\partial w} = -2(y_i - \hat{y}_i) \mathbf{X}_i + 2\lambda \text{sign}(w) \quad ; 
    \frac{\partial \text{MSE}_i}{\partial b} = -2(y_i - \hat{y}_i)
    \end{aligned}
\end{equation}
}

{
\scriptsize
\begin{equation}
    \label{eq:mse_ridge}
    \text{MSE}_i = (y_i - \hat{y}_i)^2 + \lambda  \sum_{j=1}^{m} w_j^2 = (y_i - \hat{y}_i)^2 + \lambda \|\mathbf{w}\|^2
\end{equation}
\vspace{-5.00mm}
\begin{equation}
    \label{eq:ridge_w_b}
    \begin{aligned}
    \frac{\partial \text{MSE}_i}{\partial \mathbf{w}} = -2(y_i - \hat{y}_i) \mathbf{X}_i + 2\lambda \mathbf{w}  \quad ; 
    \frac{\partial \text{MSE}_i}{\partial b} = -2(y_i - \hat{y}_i)    
    \end{aligned}
\end{equation}
}



\subsection{Online Regression using Mini-Batch Gradient Descent}
The Mini-Batch Gradient Descent (MBGD) algorithm combines the strengths of both batch gradient descent (GD) and stochastic gradient descent (SGD) by updating the model based on mini-batches\cite{ruder2016overview}. This approach strikes a balance between the robustness of GD and the computational efficiency of SGD \cite{haji2021comparison}. Utilizing mini-batches is a common technique to expedite convergence \cite{danner2015fully}. The choice of the batch size, denoted as $K$, plays a crucial role in optimizing computational resources while achieving accurate results \cite{avrithis2021iterative}. However, similar to SGD, MBGD lacks a mechanism to forget the data it has encountered. Selecting an appropriate batch size is essential; a small batch size can introduce high variance in parameter updates, while a large batch size may reduce the algorithm's stochasticity and hinder generalization \cite{keskar2016large} \cite{ruder2016overview}. The equations \ref{eq:mini-batch_mse} and \ref{eq:mini-batch-gradient-w-b} demonstrate the distinctions from SGD.

{
\scriptsize
\begin{equation}
\label{eq:mini-batch_mse}
\text{MSE}_i = \frac{1}{K}\sum_{i=1}^{K}(y_i - \hat{y}_i)^2    
\end{equation}
\vspace{-4.0mm}
\begin{equation}
\label{eq:mini-batch-gradient-w-b}
\begin{aligned}
\frac{\partial \text{MSE}_i}{\partial w} = \frac{-2}{K} \sum_{i=1}^{K} (y_i - \hat{y}_i) \cdot \mathbf{X}_i   \quad ;
\frac{\partial \text{MSE}_i}{\partial b} = \frac{-2}{K} \sum_{i=1}^{K} (y_i - \hat{y}_i) 
\end{aligned}
\end{equation}
}

\subsection{Widrow-Hoff (LMS)}
The Least Mean Squares (LMS) algorithm, introduced by Widrow and Hoff in 1960 \cite{widrow1960adaptive}, is an adaptive algorithm that employs a stochastic gradient-based approach derived from the steepest descent method. Its relative simplicity is a notable advantage, as it avoids correlation function calculations and matrix inversions, distinguishing it from other algorithms \cite{fontenla2013online}. LMS objective is to minimize the current MSE. Hence, it doesn't consider old data. However, its lack of memory may limit its ability to capture long-term dependencies or complex patterns in the data \cite{fontenla2013online}. Comparatively, the convergence of LMS can be slower than more advanced algorithms like Recursive Least Square (RLS) \cite{hsia1983convergence}. LMS relies on a constant learning rate, needing prior knowledge of input signal statistics \cite{kwong1992variable}. The LMS update rule can be expressed as shown in Eq. \ref{eq:widrow_hoff_update_rule}.

{
\scriptsize
\begin{equation}
    \label{eq:widrow_hoff_update_rule}
    w_{t+1} = w_t + 2 \eta ( w_t \cdot x_t - y_t ) x_t \quad \triangleright \eta \text{ (learning rate)} > 0
\end{equation}
}

\subsection{Recursive Least-Squares (RLS)}
RLS algorithm has infinite memory, considering ``cumulative error'' up to the current data point, unlike LMS, leading to better long-term dependency capture and faster convergence, and typically yields smaller errors compared to LMS \cite{fontenla2013online}. RLS ensures convergence to the optimal solution, equivalent to regular least squares update, but without requiring matrix inversion \cite{bittanti1988deterministic}\cite{fontenla2013online}. However, this improved performance comes at the cost of increased computational effort per iteration \cite{haykin2013adaptive}. The absence of a learning rate is a notable characteristic of this algorithm \cite{haykin2013adaptive}\cite{fontenla2013online}. The forgetting factor, represented by the symbol $\lambda$, allows the algorithm to ``forget'' or give less importance to previous data when there are changes in the target functions \cite{bittanti1988deterministic} \cite{fontenla2013online}. However, the algorithm has a potential issue known as ``numerical instability'' under finite word-length conditions due to ill-conditioning, causing inaccuracies and instability in estimated values. Techniques like data scaling and numerical stabilization are used to mitigate this issue \cite{douglas2000numerically}. RLS updates its parameters in each iteration using equations \ref{eq:rls_loss} and \ref{eq:rls_update_rule}, where $\lambda \in (0, 1]$ is a forgetting factor, and for the initial step of the algorithm W$_{0} = 0$, and P$_{0} = \delta I$, while $\delta >> 1$, $d_t$ = $y_t$ \cite{fontenla2013online}.

{
\scriptsize
\begin{equation}
\label{eq:rls_loss}
\textbf{P}_{t+1} = \frac{1}{\lambda} \left( \textbf{P}_{t} - \frac{\textbf{P}_{t} \textbf{x}_{t+1} \textbf{x}_{t+1}^\intercal \textbf{P}_{t}}{\lambda + \textbf{x}_{t+1}^\intercal \textbf{P}_{t} \textbf{x}_{t+1}} \right)    
\end{equation} 
\vspace{-3.0mm}
\begin{equation}
\label{eq:rls_update_rule}
\textbf{w}_{t+1} = \textbf{w}_{t} + \textbf{P}_{t+1} \textbf{x}_{t+1} \left( d_{t+1} - \textbf{x}_{t+1}^\intercal \textbf{w}_{t} \right)
\end{equation}  
}

\subsection{Online Passive-Aggressive (PA)}
The online passive-aggressive algorithm \cite{crammer2006online} is a group of algorithms that shares a close relationship with Support Vector Machine (SVM) methods. The name ``passive-aggressive'' is used to convey the concept that learning occurs only under specific conditions (aggressive phase), while in other situations, the model remains ``$\mathrm{passive}$'', in other words, whenever the loss is zero, the model remains unchanged even when new learning data is introduced (passive phase). PA algorithm exhibits low time complexity, offers fast convergence and efficient performance. The hyper-parameter $C$ strongly affects performance and convergence. A small $C$ leads to slower progress, while a large $C$ causes faster error reduction but increases the risk of overfitting \cite{crammer2006online}. PA is a margin-based approach which effectively handles noisy data by focusing on instances near the decision boundary \cite{crammer2006online}. PA employs the \(\epsilon\)-insensitive hinge loss function as defined in Equation \ref{eqn:pa_eps_hinge_loss}.\cite{crammer2006online}:

{
\scriptsize
\begin{equation}
    \label{eqn:pa_eps_hinge_loss}
    \text{{loss}}(w_t \cdot x_t, y_t) = \begin{cases}
0 & \text{{if }} |w_t \cdot x_t - y_t| \leq \epsilon \\
|w_t \cdot x_t - y_t| - \epsilon & \text{{otherwise}}
\end{cases}
\end{equation}
}
The parameter \(\epsilon\) is a positive user-defined value. The $\epsilon-$region is of width 2$\epsilon$, usually called hyper-slab\cite{crammer2006online} determines the sensitivity to prediction errors. The update equation for the weights, denoted as equation \ref{eqn:pa-update-rule}, encompasses three variations \cite{crammer2006online}. \(\ell_t\) is the $\epsilon-$insensitive hinge loss. 
$\tau_t \geq 0 $ is a Lagrange multiplier, in which 
$C$ is a positive parameter ``aggressiveness parameter'' whereas larger values of $C$ imply more aggressive update step.

{
\scriptsize
\begin{equation}
\label{eqn:pa-update-rule}
\begin{aligned}
\mathbf{w}_{t+1} &= \mathbf{w}_t + \text{sign}(y_t - \hat{y}_t) \tau_t \mathbf{x}_t, \\
\textrm{where} \quad \tau_t &= 
\begin{cases}
\frac{l_t}{\|\mathbf{x}_t\|^2} & \quad \text{PA-}\uppercase\expandafter{\romannumeral1} \\
\min(C, \frac{l_t}{\|\mathbf{x}_t\|^2}) & \quad \text{PA-}\uppercase\expandafter{\romannumeral2} \\
\frac{l_t}{\|\mathbf{x}_t\|^2 + \frac{1}{2C}} & \quad \text{PA-}\uppercase\expandafter{\romannumeral3}
\end{cases}
\end{aligned}
\end{equation}
}
The three variations differ in calculating the Lagrange parameter (learning rate) $\tau_t$. $PA-\uppercase\expandafter{\romannumeral1}$ sets the learning rate based solely on the loss and feature vector's squared norm. $PA-\uppercase\expandafter{\romannumeral2}$ introduces a regularization term by limiting the learning rate to a maximum value of C, preventing excessive weight updates. $PA-\uppercase\expandafter{\romannumeral3}$ combines regularization and constraint, stabilizing the updates with an additional term in the denominator \cite{crammer2006online}.

\section{OLR-WA Method}
OLR-WA combines incoming data with a pre-existing base model through a dynamic merging process. At its core, OLR-WA incorporates two vital components: 
the initial base model created from the first batch through linear regression and the incremental model, built iteratively as new data arrives.

The base model is continuously refined as new data streams in. By harnessing user-defined weights as hyper-parameters W$_{\text{base}}$ and W$_{\text{inc}}$, the model gracefully adapts to changing patterns within the data, ensuring accurate and up-to-date predictions. 

Specifically, OLR-WA calculates the weighted average of the base model and incremental models as an Exponentially Weighted Moving Average (EWMA)  \cite{shumway2000time}. By default, equal weights are assigned to both the base model and the incremental one for an iteration. However, users can adjust the weights to influence the model's adaptation rate or resilience to transient changes. This flexibility allows users to lean results toward either old or new data based on their knowledge and requirements. The weighted average equation, follows a consistent pattern regardless of the dimensionality, presented in equation \ref{eq:olr-wa-gen-equation}. This equation is used to compute the weighted average vector out the two normalized norm vectors of the base and the incremental planes, and the two user defined weights $W_{base}$, and $W_{inc}$.

\begin{figure}[t]
\centering
\includegraphics[width=.22\textwidth,height=0.06\textheight]{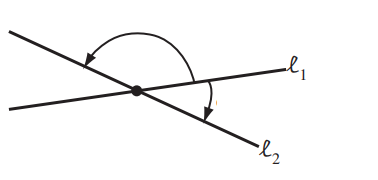}
\caption{Two Intersection Sides}
\label{fig:two_intersection_sides}
\end{figure}

{
\scriptsize
\begin{equation}
\label{eq:olr-wa-gen-equation}
    V_{\text{Avg}} = \frac{{(W_{\text{base}} \cdot V_{\text{base}} + W_{\text{inc}} \cdot V_{\text{inc}})}}{{(W_{\text{base}} + W_{\text{inc}})}}
\end{equation}
}
In practical terms, OLR-WA computes two weighted average vectors from the base and incremental models, presented as $V_{avg1}$ and $V_{avg2}$ in Algorithm \ref{alg:olrwaalg}. That can be explained as any two intersecting planes have two sides of intersection, as described in Figure \ref{fig:two_intersection_sides}.


Theoretically, to define a new hyperplane, two components are needed: a point in the plane and a vector orthogonal to the plane. The point used to define the new hyperplane is found in the intersection of the two planes, and the orthogonal vector will be $V_{avg}$. Since we defined two such average vectors corresponding to the two ``sides'' of the intersection, $V_{avg1}$, and $V_{avg2}$, the process entails defining two new weighted average planes from these inputs. OLR-WA subsequently selects the best fit. Since only the incremental data is available for evaluation, OLR-WA generates a sample of the prior data using the existing base model. This sample encompasses a comprehensive dataset. The model selects the best fit by assessing the mean square error of the whole sample. Algorithm ~\ref{alg:olrwaalg} shows the steps of OLR-WA.

Several cases arise when finding a point in the intersection of the two planes. If the planes are not parallel, a point in the intersection can be trivially obtained by solving a system of linear equations that yields a single solution in two dimensions and infinite ones in higher dimensions. In the infinite case, an arbitrary one is selected. Conversely, when the planes are parallel, two possibilities arise, as described by Layton et al. \cite{layton2014numerical}. Firstly, the ``Coincide'' scenario occurs when either the base plane or the incremental plane is precisely aligned atop the other, resulting in an infinite number of intersection points. In such straightforward instances, OLR-WA continues to the next incremental mini-batch, as no update is required using the current one, obviating the need for further algorithmic steps, as the current coefficients fulfill the objective. Secondly, the ``Parallel'' case arises when the base plane is parallel to the incremental plane, leading to the absence of an intersection point. Under this circumstance, we compute a weighted midpoint between the two models. This midpoint's position is influenced by the higher weight between W$_{\text{base}}$ and W$_{\text{inc}}$, in case of weight disparity, or it settles at the exact midpoint when the weights are equal. Nevertheless, it is crucial to note that these two cases are exceedingly rare and have not been encountered in our experiments with datasets.

In the realm of N-dimensional geometry, the definition of a new hyperplane necessitates the utilization of a norm vector and a point \cite{stewart2020calculus}. Let \(\mathbf{n} = \langle n_1, n_2, \ldots, n_N \rangle\) denote the norm vector, representing the directional characteristics of the desired hyperplane, and let \(\mathbf{P} = (X_{1_p}, X_{2_p}, \ldots, X_{N_p})\) be a point lying within the hyperplane. To establish the equation of the new hyperplane, consider an arbitrary point \(\mathbf{Q} = (x_1, x_2, \ldots, x_N)\) on the hyperplane. The vector connecting \(\mathbf{P}\) and \(\mathbf{Q}\), denoted as \(\overrightarrow{\text{PQ}} = \langle x_1 - X_{1_p}, x_2 - X_{2_p}, \ldots, x_N - X_{N_p} \rangle\), lies within the hyperplane and is orthogonal to the norm vector \(\mathbf{n}\). Hence, the orthogonality condition can be expressed as \(\mathbf{n} \cdot \overrightarrow{\text{PQ}} = n_1(x_1 - X_{1_p}) + n_2(x_2 - X_{2_p}) + \ldots + n_N(x_N - X_{N_p}) = 0\). This methodology enables the precise definition of a new hyperplane in N-dimensional space based on its norm vector and a known point lying within the hyperplane.

\begin{algorithm}
\caption{OLR-WA}
\label{alg:olrwaalg}
\footnotesize
\begin{algorithmic}[1]
\State $\text{coeff-regression$_{base}$} = \text{pseudo-inverse}(\text{X$_{base}$}, \text{y$_{base}$})$
\For{$t \gets 1$ to $T$}
\State $\text{coeff-regression$_{inc}$} = \text{pseudo-inverse}(\text{X$_{inc}$}, \text{y$_{inc}$})$
\If{coincide ($\text{coeff-regression}_{\text{base}}, 
\text{coeff-regression}_{\text{inc}}$)}
\State \textbf{continue}
\EndIf
\State 
$\text{intersection-point} = \text{get-intersection-point}(\text{coeff-regression}_{\text{base}},$ \\ \hskip\algorithmicindent\hskip\algorithmicindent 
$\text{coeff-regression}_{\text{inc}})$
\If{No intersection-point (parallel)}
    \State $\text{intersection-point} =$  $\text{weighted-mid-point}(\text{coeff-regression$_{base}$},$ \par
    \hskip\algorithmicindent\hskip\algorithmicindent $\text{coeff-regression$_{inc}$}, \text{W$_{base}$}, \text{W$_{inc}$})$
\EndIf
\State $\mathbf{V_{\text{Avg1}}} = \frac{\mathbf{(W_{\text{base}} \hspace{3pt} \cdot \hspace{3pt} \mathbf{V_{\text{base}}} \hspace{4pt}+ \hspace{4pt}\mathbf{W_{\text{inc}}} \hspace{3pt} \cdot \hspace{3pt} \mathbf{V_{\text{inc}}})}}{\mathbf{(W_{\text{base}} + \mathbf{W_{\text{inc}}})}}$
\State $\mathbf{V_{\text{Avg2}}} = \frac{\mathbf{(- W_{\text{base}} \hspace{3pt} \cdot \hspace{3pt}  \mathbf{V_{\text{base}}} \hspace{4pt}+ \hspace{4pt} \mathbf{W_{\text{inc}}} \hspace{3pt} \cdot \hspace{3pt} \mathbf{V_{\text{inc}}})}}{\mathbf{(W_{\text{base}} + \mathbf{W_{\text{inc}}})}}$
\State $\text{space-coeff-1} = \text{define-new-hyperplane}(\mathbf{V_{\text{Avg1}}}, \text{intersection-point})$
\State $\text{space-coeff-2} = \text{define-new-hyperplane}(\mathbf{V_{\text{Avg2}}}, \text{intersection-point})$
\State \text{X$_{combined}$}, \text{y$_{combined}$} = sample-and-combine($\text{X$_{inc}$}$, $\text{y$_{inc}$}$, \\ \hskip\algorithmicindent\hskip\algorithmicindent $\text{coeff-regression$_{base}$}$)
\State $\text{err$_{v1}$} = \text{MSE}(\text{space-coeff-1}, \text{X$_{combined}$}, \text{y$_{combined}$})$
\State $\text{err$_{v2}$} = \text{MSE}(\text{space-coeff-2}, \text{X$_{combined}$}, \text{y$_{combined}$})$
\If{$\text{err$_{v1}$} < \text{err$_{v2}$}$}
    \State $\text{coeff-regression$_{base}$} \gets \text{space-coeff-1}$
\Else
    \State $\text{coeff-regression$_{base}$} \gets \text{space-coeff-2}$
\EndIf
\EndFor
\State \textbf{return} $\text{coeff-regression$_{base}$}$
\end{algorithmic}
\end{algorithm}

OLR-WA utilizes pseudo-inverse linear regression per iteration. Consequently, the time complexity of the pseudo-inverse is dominated by the $M \times N$ matrix multiplied by $N \times M$ matrix, with a time complexity of $O(NM^2)$, where $N$ is the number of samples and $M$ the number of features. OLR-WA processes data incrementally, enabling efficient computation by handling smaller data batches, compared to the batch version, OLR-WA's time complexity is estimated to be approximately $O(KM^2)$ for each iteration, where $K$ represents the number of samples in the mini-batch. This analysis presents the complexity in a ``static context''; however, it is important to consider the practical implications. In practice, the utilization of pseudo-inverse as a batch model would require reprocessing all existing and new data points if encountering new data, resulting in a significantly higher time complexity compared to online models. This is primarily because the batch model's computational burden increases with the growing size of data points ($N$) when new data is introduced, while, OLR-WA as an online model, $K$ remains fixed as a small mini-batch throughout.

\subsection{Hyperparameters Tuning}
OLR-WA involves four hyperparameters, namely W$_{\text{base}}$, W$_{\text{inc}}$, K- mini-batch size, and BK- base model size. This section explores various alternatives for tuning these hyperparameters effectively.
\begin{enumerate}
    \item \small BK - represents the base model size, i.e., the number of data points of the base model. The analysis of BK size reveals that it does not significantly impact performance. Surprisingly, the model exhibits robust performance even with as few as 1\% to 10\% of the total training data points.
     \item \small K - refers to the mini-batch size, i.e., the size of incremental mini-batches utilized during processing. As we utilize pseudo-inverse, K is constrained to be $\geq M$. OLR-WA constraint can be represented using equation \ref{eq:olr_wa_batch_size}, where $U$ denotes the user-defined batch size, $M$ represents the number of dimensions, and $\mathbb{Z}$ signifies a natural number $\geq$ 1. Based on our experimental findings, we recommend selecting $\mathbb{Z}^{+} \geq 4$ would yield results nearly identical to the standard batch regression version. 
    \item \small W$_{\text{base}}$, and W$_{\text{inc}}$ -  the user-defined weights W$_{\text{base}}$ and W$_{\text{inc}}$ are scalar values, It is not mandatory for them to add up to 1 since OLR-WA automatically scales them, they offer various alternatives for assigning weights based on the desired scenario. The following options are available:
    \begin{enumerate}
        \item \small Assigning equal weights: Set W$_{\text{base}} = 0.5$ and W$_{\text{inc}} = 0.5$. This default option performs effectively in all scenarios. It is important to consider that this choice assigns equal weights to both the base model, which may represent thousands of past data points, and the incremental model, which may represent only a few data points. However, this default setting guarantees OLR-WA to converge, as recent data points are given more weight and older data points are given exponentially decreasing weights, which yields the utilization of Exponentially Weighted Moving Average (EWMV), which is mathematically proven to converge to new patterns. 
        
    \item \small Dynamic weight assignment: Employ dynamic weights based on the number of accumulative data points and incremental points. In other words, if the base model represents 1,000 data points, while the incremental model represents 10 points, set W$_{\text{base}} = 1$ and W$_{\text{inc}} = 0.01$. An example of where this might fit is a machine translation system since both the previously trained model and the additional training data have equal importance in generating accurate translations.
    \item \small Time-Based Data \cite{cormode2009forward}, this scenario is used when the user aims to expedite the convergence to new patterns, requires giving higher weight to new data points, allowing the model to adapt to emerging data patterns while gradually forgetting old data. For instance, users may define values like W$_{\text{inc}} = 2$ and W$_{\text{base}} = 0.1$, wherein W$_{\text{inc}}$ is 20 times higher than W$_{\text{base}}$.
    \item \small Confidence-Based Data \cite{prasad2008decision}, in this scenario, higher weight is assigned to old data points, as the constructed existing model represents a trusted predictor. Slight model updates are allowed as new data streams in. For example, users may define values like W$_{\text{inc}} = 0.1$ and W$_{\text{base}} = 2$, wherein W$_{\text{base}}$ is 20 times higher than W$_{\text{inc}}$. This setting might be useful in the presence of cases like trusted sources, historical data quality, and samples that are certified by experts to be more accurate and reliable.
    \end{enumerate}    
\end{enumerate}
\begin{equation}
\footnotesize
\label{eq:olr_wa_batch_size}
\text{K (Mini-Batch Size)} = \max\left(U, (M \times \mathbb{Z}^{+})\right)
\end{equation}

\section{Experiments}
In this section, we conducted experiments using synthetic and real public datasets with a rigorous methodology for robustness. We employed 5 random seeds for seed averaging and 5-fold cross-validation for comprehensive evaluation. Each reported $r^2$ value underwent validation through 25 experiments with different seeds and data splits, ensuring stability and consistency in results, considering variations from seed initialization and data partitions.

In our experiments, we followed strict guidelines to ensure impartial model evaluation and prevent bias.

\begin{enumerate}
    \item \small Initialization - During the initialization phase, all models' weights were set to an array of zeros, establishing a uniform and consistent starting point for each model.
     \item \small Dataset Selection - Utilizing 14 diverse datasets (see table \ref{tab:datasets-properties}), our approach covered a broad range of situations. We included synthetic and public datasets for linear regression analysis, spanning low to high dimensions and different sample sizes. Adversarial datasets were also integrated for Time-Based and Confidence-Based scenarios.
    \item \small Feature Engineering - We applied minimal feature engineering techniques like normalization and one-hot encoding to preserve data integrity and maintain balanced model evaluation, exercising caution to avoid undue impact from feature manipulation.
    \item \small Hyper-parameter Tuning - We fine-tuned hyper-parameters for each model, considering factors like dataset size and feature count. Optimal outcomes were identified and reported. Caution was exercised in adjusting parameters like epochs to prevent excessively long execution times while measuring experiment execution time.
\end{enumerate}

\subsection{Performance Analysis on Normal Linear Regression Scenarios}
Table \ref{tab:datasets-properties} (DS1 to CCPP) 
presents the key properties of the specific datasets employed in this study. 

\textit{Performance Evaluation }
Table \ref{tab:algorithm-performance} presents the performance results for each dataset, along with the hyper-parameters utilized for each model. The inclusion of these hyper-parameters enhances transparency and facilitates a comprehensive analysis of the models' performance. Below are some observed insights about the performance evaluation for this experiment:
\begin{enumerate}
    \item \small The batch model, included as a benchmark for evaluating accuracy, achieved the highest precision across all datasets.
    \item \small OLR-WA performance is remarkably significant and very close to the batch model. The difference is very slight and almost in the third digit after the decimal point across all datasets, except the 1KC dataset, where the batch model performance is 0.93615 while OLR-WA is 0.90773.
    \item \small RLS performance significantly deteriorates on the more challenging DS4 dataset, where it achieves an $r^2$ value of approximately 0.62667. Additionally, it demonstrates a low $r^2$ score of 0.66197 on the CCPP dataset.
    \item \small PA has demonstrated its proficiency in capturing underlying patterns and relationships effectively, as evidenced by competitive $r^2$ values across various datasets. However, its performance on the 1KC dataset appears to be somewhat limited, with an $r^2$ value of approximately 0.78565, in contrast to the batch model's 0.93615. Similarly, on the CCPP dataset, PA achieved an $r^2$ value of 0.66327, notably lower than the batch model's 0.92855. 
    \item \small The Widrow-Hoff algorithm exhibits good performance in most of the datasets, although it experienced some degradation with an $r^2$ value of 0.60772 in the 1KC dataset, whereas the batch model $r^2$ value is 0.93615.
\end{enumerate}
In conclusion, our focus on OLR-WA resulted in consistent top-tier performance across all datasets.

\begin{table*}[t]
\captionsetup{justification=centering}
\caption{1st Experiment: Performance Analysis on Normal Regression Scenarios\\
{\scriptsize Summary of Performance Measures using R-Squared on the Last Iteration; \quad  
\scriptsize E: epochs, $\mathbb{Z}^{+}$: Natural Number $\geq$ 1}}
\centering
\renewcommand{\theadalign}{cc} 
\renewcommand{\theadfont}{\small} 
\setlength{\tabcolsep}{13pt} 
\begin{adjustbox}{width=\textwidth} 
\begin{tabular}{|c|*{9}{c|}}
\hline
\multirow{2}{*}{\thead{Datasets}} & \multicolumn{9}{c|}{\thead{Algorithms}} \\ \cline{2-10} &
								 \rotatebox[origin=c]{90}{\thead{\textbf{Batch Regression} \\(Pseudo-Inverse)}} & 
                                 \rotatebox[origin=c]{90}{\thead{\textbf{SGD}\\ {\scriptsize $\eta=0.01$ [DS3, DS3 =0.01},\\ \scriptsize{ KCHSD 0.0001]}, \\ {\scriptsize E=N $\times$ $(\mathbb{Z}^{+}=2)$ [1KC $\mathbb{Z}^{+}=3$,}\\{\scriptsize CCPP $\mathbb{Z}^{+}=5$]}}} & 
                                 \rotatebox[origin=c]{90}{\thead{\textbf{MBGD} \\ {\scriptsize $\eta=0.01$ [DS4= 0.001],}\\{\scriptsize K=M $\times (\mathbb{Z}^{+} = 5)$,}\\{\scriptsize $E=\frac{N}{K} \times (\mathbb{Z}^{+} =10)$ [DS1 $\mathbb{Z}^{+}$=5,}\\{\scriptsize DS3 $\mathbb{Z}^{+}$=100, KCHSD $\mathbb{Z}^{+}$=20,}\\{\scriptsize 1KC, CCPP $\mathbb{Z}^{+}$=40]}}} & 
                                 \rotatebox[origin=c]{90}{\thead{\textbf{LMS}\\ {\scriptsize $\eta=0.01$[1KC=0.001},\\ \scriptsize{KCHSD=0.0001]}}} & 
                                 \rotatebox[origin=c]{90}{\thead{\textbf{ORR}\\ {\scriptsize $\eta=0.01$ [DS3, DS4, KCHSD,}\\ \scriptsize{MCPD=0.001], E=N $\times (\mathbb{Z}^{+}=2)$}\\{\scriptsize [1KC $\mathbb{Z}^{+}$=3, KCHSD, CCPP}\\ \scriptsize{ $\mathbb{Z}^{+}=5$], $\lambda=0.1$ [1KC,}\\{\scriptsize [MCPD,KCHSD,CCPP=0.001]}}} & 
                                 \rotatebox[origin=c]{90}{\thead{\textbf{OLR}\\ {\scriptsize $\eta=0.01$ [DS3, DS4, MCPD, KCHSD}\\{\scriptsize = 0.001], E=N $\times (\mathbb{Z}^{+}=2)$}\\{\scriptsize [1KC $\mathbb{Z}^{+}$=3, KCHSD, CCPP $\mathbb{Z}^{+}=5$], }\\{\scriptsize $\lambda=0.1$ [MCPD, CCPP=0.01],}\\ \scriptsize{[1KC, KCHSD=$0.001$],[KCHSD=$0.0001$]}}} & 
                                 \rotatebox[origin=c]{90}{\thead{\textbf{RLS} \\ {\scriptsize $\lambda=.99$, $\delta=0.01$}}} &                        
                                 \rotatebox[origin=c]{90}{\thead{\textbf{PA}\\ 
                                 {\scriptsize \text{PA-}\uppercase\expandafter{\romannumeral3}
                                 }\\
                                 {\scriptsize $C=0.01$, $\epsilon=0.01$},\\ \scriptsize{ [DS1, DS2, DS3}\\ \scriptsize{$C=0.1$, $\epsilon=0.1$]}}} & 
                                 \rotatebox[origin=c]{90}{\thead{\textbf{OLR-WA}\\ {\scriptsize W$_{base}=.5$, W$_{inc}=.5$}\\{\scriptsize [KCHSD W$_{base}=.9$, W$_{inc}=.1$],}\\{\scriptsize $BK=N \times (\mathbb{Z}^{+} = 0.1)$,}\\{\scriptsize K=M $\times$ $(\mathbb{Z}^{+} = 5)$}}} \\ \hline
\thead{DS1 \cite{shaira2023_olr_wa_synth_datasets}}&\makecell{0.97637}& \makecell{0.97561} & \makecell{0.97580} & \makecell{0.97616} & \makecell{0.96760}& \makecell{0.97565}& \makecell{0.97624} & \makecell{0.97378} & \makecell{0.97423}   \\ \hline
\thead{DS2 \cite{shaira2023_olr_wa_synth_datasets}}& \makecell{0.98418} & \makecell{0.98231}& \makecell{0.98384 }& \makecell{0.98243}& \makecell{0.96987}& \makecell{0.97959}& \makecell{0.98268}&  \makecell{0.97642}& \makecell{0.98260}  \\ \hline
\thead{DS3 \cite{shaira2023_olr_wa_synth_datasets}}& \makecell{0.98299} & \makecell{0.98116 }& \makecell{0.98290}& \makecell{0.98161 }& \makecell{0.96901 }& \makecell{0.97933}& \makecell{0.96001 }& \makecell{0.96705 }& \makecell{0.98204}  \\ \hline
\thead{DS4 \cite{shaira2023_olr_wa_synth_datasets}}& \makecell{0.92973 } & \makecell{0.90663    }& \makecell{0.92964 }& \makecell{0.90715 }& \makecell{0.85535 }& \makecell{0.86412 }& \makecell{0.62667 }& \makecell{0.87255  }& \makecell{0.92464 }  \\ \hline
\thead{MCPD \cite{MCPD_DS}}& \makecell{ 0.74321 } & \makecell{0.73092 }& \makecell{0.73007 }& \makecell{0.73637  }& \makecell{0.74070 }& \makecell{0.74217 }& \makecell{0.73830 }&  \makecell{0.72893 } & \makecell{0.74080 }  \\ \hline
\thead{1KC\cite{1KC_DS}}& \makecell{0.93615 } & \makecell{0.92930 }& \makecell{0.91195  }& \makecell{0.60772  }& \makecell{0.90294 }& \makecell{0.90322 }& \makecell{0.85503  }& \makecell{0.78565  }& \makecell{0.90773 }  \\ \hline
\thead{KCHSD \cite{KCHS_DS}}& \makecell{0.57859}  &\makecell{0.57277 } &\makecell{0.57431 } & \makecell{0.56344 }& \makecell{0.56432  }& \makecell{0.57040}& \makecell{0.48088}&  \makecell{0.53533  }& \makecell{0.57395 }  \\ \hline
\thead{CCPP \cite{CCPP_DS}}&  \makecell{0.92855}&\makecell{0.91838} & \makecell{0.91989}& \makecell{0.89785}& \makecell{0.92550}& \makecell{0.92611}&\makecell{0.66197}& \makecell{0.66327}& \makecell{0.92202}  \\ \hline
\end{tabular}
\end{adjustbox}
\label{tab:algorithm-performance}
\end{table*}

\subsection{Performance Analysis on Adversarial Scenarios}

We employed the datasets summarized in Table \ref{tab:datasets-properties} (DS5 to DS8). These datasets are characterized by adversarial scenarios, where the model initially follows a specific distribution and correlation but undergoes a significant shift toward the opposite direction at a certain point. Experiments using the first two datasets, namely DS5 and DS6, were designed to assess the model's performance on time-based scenarios. This method acknowledges that recent data points are more likely to be relevant to the current situation and gives them more importance. In other words, the model is assessed on its ability to follow the new data distribution. On the other hand, the subsequent experiments that use the rest of the datasets, DS7 and DS8, were designed to favor the inclusion of old data, allowing us to evaluate the model's capability in achieving confidence-based scenarios in which older data points are considered more accurate or reliable. The model's performance was evaluated based on its capacity to maintain alignment with the historical data, which signifies our confidence in its accuracy.

\begin{table*}[t] 
\centering
\captionsetup{justification=centering}
\caption{2nd Experiment: Performance Analysis on Adversarial Scenarios\\
{\scriptsize  Summary of Performance Measures using R-Squared on the Last Iteration; \quad   \vspace{-.06in}
\scriptsize N/A: Minus R-squared, E=Epochs, $\mathbb{Z}^{+}$: Natural Number $\geq$ 1}}
\renewcommand{\theadalign}{cc} 
\renewcommand{\theadfont}{\small} 
\setlength{\tabcolsep}{13pt} 
\begin{adjustbox}{width=\textwidth} 
\begin{tabular}{|c|*{9}{c|}}
\hline
\multirow{2}{*}{\thead{Datasets}} & \multicolumn{9}{c|}{\thead{Algorithms}} \\ \cline{2-10} &
								 \rotatebox[origin=c]{90}{\thead{\textbf{Batch Regression} \\(Pseudo-Inverse)}} & 
                                 \rotatebox[origin=c]{90}{\thead{\textbf{SGD} \\ {\scriptsize $\eta=$[DS5, DS7 = 0.01,}\\{\scriptsize DS6, DS8=0.001],}\\{\scriptsize E=N $\times (\mathbb{Z}^{+} = 2)$}}} & 
                                 \rotatebox[origin=c]{90}{\thead{\textbf{MBGD} \\ {\scriptsize $\eta=0.01$, K=M $\times (\mathbb{Z}^{+} = 5)$}\\{\scriptsize E=$\frac{N}{K} \times (\mathbb{Z}^{+} = 2)$}
                                 }} & 
                                 \rotatebox[origin=c]{90}{\thead{\textbf{LMS}\\ {\scriptsize $\eta=$ [DS5, DS7 = 0.01,}\\{\scriptsize DS6, DS8 = 0.001]}}} & 
                                 \rotatebox[origin=c]{90}{\thead{\textbf{ORR}\\ {\scriptsize $\eta=$ [DS5, DS7 = 0.01,}\\{\scriptsize DS6, DS8 = 0.001]} \\{\scriptsize E = N $\times (\mathbb{Z}^{+} = 2)$,}\\{\scriptsize $\lambda = $[DS5, DS7 = 0.1,}\\{\scriptsize DS6, DS8 = 0.01]}
                                 }} & 
                                 \rotatebox[origin=c]{90}{\thead{\textbf{OLR}\\ {\scriptsize $\eta=$ [DS5, DS7 = 0.01,}\\{\scriptsize DS6, DS8 = 0.001]} \\{\scriptsize E = N $\times (\mathbb{Z}^{+} = 2)$,}\\{\scriptsize $\lambda = $[DS5, DS7 = 0.1,}\\{\scriptsize DS6, DS8 = 0.01]}
                                 }} & 
                                 \rotatebox[origin=c]{90}{\thead{\textbf{RLS} \\ {\scriptsize $\lambda$= [DS5, DS6 = .99,}\\{\scriptsize DS7, DS8 = .18], $\delta=0.01$}}} &                                 
                                 \rotatebox[origin=c]{90}{\thead{\textbf{PA}\\ 
                                 {\scriptsize \text{PA-}\uppercase\expandafter{\romannumeral3}}\\
                                 {\scriptsize $C=0.1$, $\epsilon=0.1$}\\ \scriptsize{[DS6 $C=.01$, $\epsilon=.01$]}}} & 
                                 \rotatebox[origin=c]{90}{\thead{\textbf{OLR-WA}\\ {\scriptsize W$_{base}=$ [DS5, DS6 = .1,}\\{\scriptsize DS7, DS8 = 4], W$_{inc}=$}\\{\scriptsize [DS5, DS6 = 2, DS7, DS8 =}\\{\scriptsize 0.01] $BK=N \times 0.1$,}\\{\scriptsize K=M $\times (\mathbb{Z}^{+} = 5)$}}} \\ \hline
\thead{DS5 \cite{shaira2023_olr_wa_synth_datasets}}&\makecell{N/A}& \makecell{N/A} & \makecell{N/A} & \makecell{0.98528} & \makecell{N/A}& \makecell{N/A}& \makecell{0.98546} & \makecell{0.97812} & \makecell{0.98498}   \\ \hline
\thead{DS6 \cite{shaira2023_olr_wa_synth_datasets}}& \makecell{N/A} & \makecell{N/A}& \makecell{N/A}& \makecell{0.93810}& \makecell{N/A}& \makecell{N/A}& \makecell{0.87134}& \makecell{0.91145}& \makecell{0.93634}  \\ \hline
\thead{DS7 \cite{shaira2023_olr_wa_synth_datasets}}&  \makecell{N/A} & \makecell{N/A}& \makecell{N/A}& \makecell{N/A}& \makecell{N/A}& \makecell{N/A }& \makecell{N/A}& \makecell{N/A}& \makecell{0.97815}  \\ \hline
\thead{DS8 \cite{shaira2023_olr_wa_synth_datasets}}& \makecell{N/A} & \makecell{N/A}& \makecell{N/A}& \makecell{N/A}& \makecell{N/A}& \makecell{N/A }& \makecell{N/A}& \makecell{N/A }& \makecell{0.93191}  \\ \hline
\end{tabular}
\end{adjustbox}
\label{tab:algorithm-adv-performance}
\end{table*}

\textit{Performance Evaluation }
Table \ref{tab:algorithm-adv-performance} presents the performance results for each dataset, along with the hyper-parameters utilized for each model. Based on the experimental findings of DS5 and DS6, specifically designed for Time-Based scenarios. Notably, the OLR-WA, LMS, PA, and RLS algorithms have demonstrated remarkable outcomes in terms of their ability to adapt to dynamic changes in the data. In contrast, the performance results of other online models, namely SGD, MBGD, ORR, OLR, and even the standard batch model, were notably unsatisfactory within this specific scenario. 

The experimental results obtained from DS7 and DS8, which are tailored for confidence-based scenarios, highlight the significance of the model's capability to effectively incorporate previous data for improved performance. Particularly, the OLR-WA algorithm has exhibited impressive outcomes by consistently adapting to changes in data patterns while maintaining a conservative approach. Conversely, the performance of all other online models proved to be notably inadequate within this particular scenario.

The accomplishments of the confidence-based OLR-WA algorithm hold significant relevance across various scenarios, such as sentiment analysis tasks, where certain labeled data points have undergone verification by experts or trusted sources. In such cases, assigning higher weights to these points based on their elevated confidence levels can prove advantageous. Similarly, consider the example of an Amazon store with an extensive product inventory and a daily influx of hundreds of new products. In this context, employing a confidence-based approach, which inherently favors the larger existing product pool, can potentially enhance performance.
\begin{table*}[t]
\centering
\captionsetup{justification=centering} 
\caption{3rd Experiment: Convergence Analysis \\
\scriptsize Models Performance by Number of Data Points Considered on DS9.
\scriptsize Hyperparameters are the same as Fig. \ref{fig:models_cost_convergence} }
\vspace{-2.00mm}
\adjustbox{width=\textwidth}{
\begin{tabular}{|c|*{16}{c|}}
\hline
\multirow{2}{*}{\textbf{Model}} & \multicolumn{16}{c|}{\textbf{Number of Training Data Points}} \\
\cline{2-17}
& \textbf{10} & \textbf{20} & \textbf{30} & \textbf{40} & \textbf{50} & \textbf{60} & \textbf{70} & \textbf{80} & \textbf{90} & \textbf{100} & \textbf{110} & \textbf{120} & \textbf{130} & \textbf{140} & \textbf{150} & \textbf{Final}\\
\hline
\textbf{OLR-WA}         &0.86265&0.87026&0.87771&0.88329&0.88861&0.89288&0.89591&0.89875&0.90098&0.90161&0.90433&0.90615&0.90810&0.90986&0.91084&0.91293\\
\hline
\textbf{SGD}    &0.32402&0.50374&0.62607&0.72745&0.78288&0.82651&0.85521&0.87468&0.88738&0.89584&0.90132&0.90371&0.90599&0.90806&0.90799&0.91233\\
\hline
\textbf{MBGD}
&0.03395&0.06898&0.09958&0.13078&0.15887&0.19059&0.21955&0.24971&0.27298&0.29824&0.31832&0.34179&0.36481&0.38586&0.40658&0.91505\\
\hline
\textbf{LMS} &0.25216&0.48899&0.66102&0.72896&0.78882&0.82736&0.84889&0.86801&0.88491&0.89585&0.90026&0.90347&0.90625&0.90919&0.91053&0.91348\\
\hline
\textbf{ORR}   &0.28438&0.44982&0.59813&0.69093&0.74606&0.78599&0.81481&0.84090&0.85622&0.86723&0.88113&0.88760&0.89168&0.89530&0.89552&0.90296\\
\hline
\textbf{OLR}   &0.27546&0.51746&0.64119&0.72773&0.77723&0.81353&0.84392&0.86510&0.88134&0.89082&0.89761&0.90173&0.90476&0.90693&0.90769&0.91299\\
\hline
\textbf{RLS}     &0.13441&0.30085&0.43636&0.49871&0.57590&0.63265&0.67494&0.71707&0.75557&0.78608&0.80571&0.82775&0.83981&0.85533&0.87163&0.91513\\
\hline
\textbf{PA}            &0.76154&0.88878&0.89449&0.90396&0.86813&0.89807&0.90726&0.91309&0.90740&0.90570&0.90363&0.90456&0.90786&0.90941&0.90577&0.90626\\
\hline
\end{tabular}
}
\label{tab:moedls-convergence-r-squared}
\end{table*}

\begin{figure*}[h]
\centering
\includegraphics[width=.9\textwidth,height=0.29\textheight]{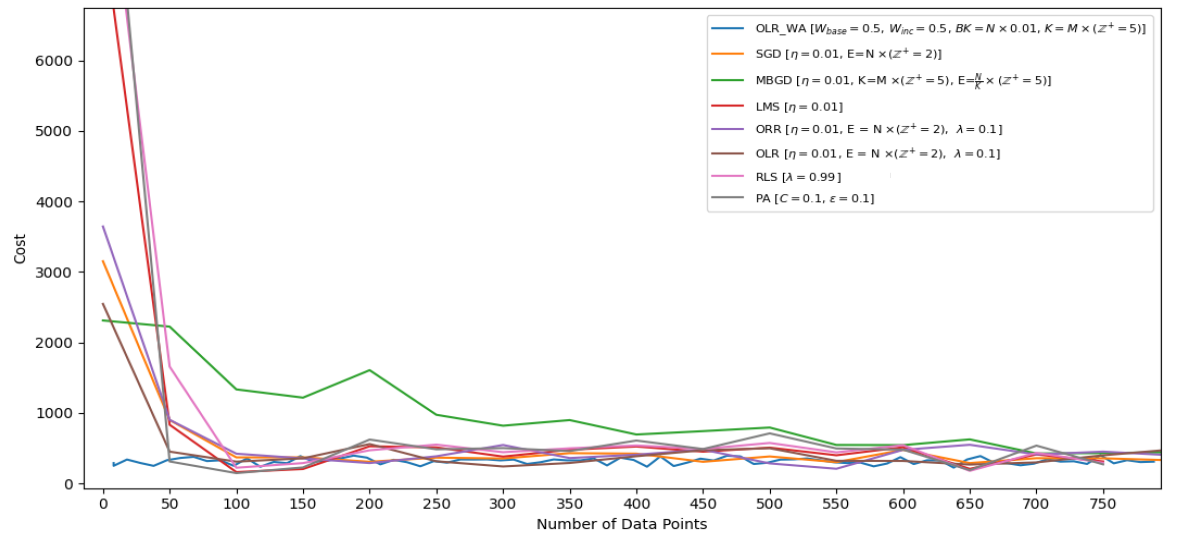}
\caption{Models Convergence on DS10}
\label{fig:models_cost_convergence}
\end{figure*}

\subsection{Convergence Analysis}
This experiment aims to conduct a comparative analysis of the convergence behavior of the aforementioned online regression models, with a specific focus on OLR-WA. By comparing its convergence performance to that of other existing online regression models, we seek to understand OLR-WA's ability to reach a stable solution. The $r^2$ will be utilized as the comparative metric, and the Mean Squared Error (MSE) will be utilized for plotting the convergence curve. Those two measures will be evaluated at each iteration. The outcomes of this analysis will contribute to a comprehensive understanding of the strengths, limitations, and potential applications of these models across various domains.

Within this experimental context, DS9 \cite{shaira2023_olr_wa_synth_datasets} and DS10 \cite{shaira2023_olr_wa_synth_datasets} serve as our reference datasets. Datasets are characterized by reduced dimensionality for better showing the convergence behavior, as with the existence of mini-batch models with higher dimensions, and consequently a higher mini-batch size, many models have already achieved convergence, which will not fit for showing the convergence in settings based on data points count.

The $r^2$ values per iteration are succinctly displayed in Table \ref{tab:moedls-convergence-r-squared}. Observations reveal that among the initial 10 data points, OLR-WA attained the highest score of 0.86265 when compared to the other models. Conversely, PA demonstrated its rapid convergence capabilities by stabilizing at a high $r^2$ value by the 70th data point. Additionally, OLR-WA exhibited sustained high $r^2$ values starting from the 90th data point. Moreover, a majority of the models displayed convergence to high $r^2$ values around the 150th data point.

The rapid convergence of OLR-WA hinges on the size of its initial mini-batch, denoted as BK. In our experiment, we intentionally constrained this size to a minimum of 1\% as an attempt to scrutinize OLR-WA's behavior under challenging conditions. However, it's noteworthy that expanding the BK size can significantly expedite the convergence of OLR-WA, enabling it to attain exceptionally high scores right from the outset. The phenomenon of OLR-WA's rapid convergence can be attributed to the utilization of batch regression (pseudo-inverse), where an increased number of samples within the mini-batch translates to a swifter acquisition of higher $r^2$ values.

Figure \ref{fig:models_cost_convergence} illustrates the convergence behavior of different algorithms, as measured by the MSE. With employing DS10 \cite{shaira2023_olr_wa_synth_datasets}, it is evident from the figure that the majority of the algorithms reach the desired performance level around epoch 150. However, it is worth noting that the OLR-WA algorithm exhibits a distinct convergence pattern, commencing with a significantly lower cost level and consistently sustaining this diminished cost throughout. This observation emphasizes the unique behavior and progression of the OLR-WA algorithm compared to the other algorithms.


\begin{table}[t]
\centering
\captionsetup{justification=centering}
\caption{Dataset Properties.\\ {\scriptsize N.R: Normal Regression, T.B: Time-Based, C.B: Confidence-Based, C: Convergence}}
\renewcommand{\theadalign}{bc} 
\renewcommand{\theadfont}{\bfseries\scriptsize} 
\setlength{\tabcolsep}{2pt} 
\begin{tabular}{cccccccc}
\toprule
\rotatebox{90}{\thead{Dataset}} & 
\rotatebox{90}{\thead{Type}} & 
\rotatebox{90}{\thead{Data-\\points}} & 
\rotatebox{90}{\thead{Dimen-\\sions}} & 
\rotatebox{90}{\thead{Noise}} & 
\rotatebox{90}{\thead{Train\\(5-Fold)}} & 
\rotatebox{90}{\thead{Test\\(5-Fold)}} & 
\rotatebox{90}{\thead{Focus}} \\
\midrule
DS1 \cite{shaira2023_olr_wa_synth_datasets} & Synth. & 1k & 3 & 10 & 800 & 200 & N.R \\
DS2 \cite{shaira2023_olr_wa_synth_datasets} & Synth. & 10k & 20 & 20 & 8k & 2k & N.R \\
DS3 \cite{shaira2023_olr_wa_synth_datasets} & Synth. & 10k & 200 & 25 & 8k & 2k & N.R \\
DS4 \cite{shaira2023_olr_wa_synth_datasets} & Synth. & 50k & 500 & 50 & 40k & 10k & N.R \\
MCPD \cite{MCPD_DS} & Real & 1.3k & 7 & N.A & 1.1k & 267 & N.R \\
1KC \cite{1KC_DS} & Real & 1k & 5 & N.A & 800 & 200 & N.R \\
KCHSD \cite{KCHS_DS} & Real & 21.6k & 21 & N.A & 17.3k & 4.3k & N.R \\
CCPP \cite{CCPP_DS} & Real & 9.6k & 5 & N.A & 7.7k & 1.9k & N.R \\
DS5 \cite{shaira2023_olr_wa_synth_datasets} & Synth. & 5k & 20 & 20 & 4k & 1k & T.B \\
DS6 \cite{shaira2023_olr_wa_synth_datasets} & Synth. & 10k & 200 & 40 & 8k & 2k & T.B \\
DS7 \cite{shaira2023_olr_wa_synth_datasets} & Synth. & 5k & 20 & 20 & 4k & 1k & C.B \\
DS8 \cite{shaira2023_olr_wa_synth_datasets} & Synth. & 10k & 200 & 40 & 8k & 2k & C.B \\
DS9 \cite{shaira2023_olr_wa_synth_datasets} & Synth. & 1k & 2 & 20 & 800 & 200 & C \\
DS10 \cite{shaira2023_olr_wa_synth_datasets} & Synth. & 1k & 2 & 40 & 800 & 200 & C \\
\bottomrule
\end{tabular}
\label{tab:datasets-properties}
\end{table}


\section{Conclusion and Future Work}

To ensure comprehensive evaluation, three experiments utilized 14 datasets with seed averaging and 5-fold cross-validation. The goal was to cover diverse situations. Results are summarized as follows: 1. OLR-WA excelled consistently in normal linear regression scenarios. 2. LMS, RLS, PA, and OLR-WA achieved notable results in time-based adversarial regression, while OLR-WA stood out in confidence-based scenarios. 3. In convergence analysis, OLR-WA demonstrated rapid high performance within the first 10 data points and sustained it from the 90th point.

This paper presents several significant contributions. 1. it introduces OLR-WA, a novel and versatile online linear regression model. 2. OLR-WA performance is comparable to the batch model, showcasing adaptability and effectiveness in real-time learning scenarios. 3. OLR-WA is capable of managing dynamic data, adjusting to emerging changes, and adhering to evolving data patterns, additionally, it is the sole model that can handle confidence-based scenarios, in which older data points are considered more accurate or reliable. 4. OLR-WA offers a level of flexibility beyond that of batch models and surpasses other online models in its ability to finely control the model's behavior through hyperparameter tuning. 5. OLR-WA exhibits exceptional performance in terms of rapid convergence, surpassing most other online models with consistently achieving high $r^2$ values from the first iteration to the last iteration, even when initialized with minimal amount of data points, as little as 1\% to 10\% of the total training data points. In conclusion, the remarkable performance of OLR-WA reinforces its adaptability and effectiveness across various contexts, establishing it as a valuable solution for tasks involving online linear regression.

This work offers promising opportunities for expansion, aiming to enrich the capacities of OLR-WA and extend its versatility, first by exploring the feasibility of implementing automatic weight selection through the preservation of multiple incremental models in memory and conducting a comparative analysis of their properties against the existing model. Second, extending the application of OLR-WA to diverse model types, such as classification models, presents an intriguing avenue for research. Additionally, a potential research direction involves a comprehensive study to assess statistical significance thoroughly. Lastly, delving into the potential utility of OLR-WA in addressing non-linear relationships holds promise as a worthy area of investigation in future research endeavors.

\bibliographystyle{IEEEtran}
\bibliography{references}
\end{document}